\documentclass[conference]{IEEEtran}

\usepackage[cmex10]{amsmath}
\usepackage{amsthm}
\usepackage{amssymb}
\usepackage{mathrsfs}
\usepackage{graphicx}
\usepackage{float}
\usepackage{array}
\usepackage{epstopdf}
\usepackage{multirow}
\usepackage{color,soul}
\usepackage{epstopdf}
\usepackage{subcaption}

\begin{document}


\title{Deep-Emotion: Facial Expression Recognition Using Attentional Convolutional Network}

\author{Shervin~Minaee$^{1}$, Amirali Abdolrashidi$^2$  \\ $^1$Expedia Group \\ $^2$University of California, Riverside}

\maketitle

\begin{abstract}
Facial expression recognition has been an active research area over the past few decades, and it is still challenging due to the high intra-class variation. 
   Traditional approaches for this problem rely on hand-crafted features such as SIFT, HOG and LBP, followed by a classifier trained on a database of images or videos. 
   Most of these works perform reasonably well on datasets of images captured in a controlled condition, but fail to perform as good on more challenging datasets with more image variation and partial faces. 
   In recent years, several works proposed an end-to-end framework for facial expression recognition, using deep learning models. 
   Despite the better performance of these works, there still seems to be a great room for improvement.
   In this work, we propose a deep learning approach based on attentional convolutional network, which is able to focus on important parts of the face, and achieves significant improvement over previous models on multiple datasets, including FER-2013, CK+, FERG, and JAFFE.
   We also use a visualization technique which is able to find important face regions for detecting different emotions, based on the classifier's output. 
  Through experimental results, we show that different emotions seems to be sensitive to different parts of the face.
\end{abstract}

\IEEEpeerreviewmaketitle

\section{Introduction}
Emotions are an inevitable portion of any inter-personal communication. They can be expressed in many different forms which may or may not be observed with the naked eye. 
Therefore, with the right tools, any indications preceding or following them can be subject to detection and recognition.
There has been an increase in the need to detect a person's emotions in the past few years. 
There has been interest in human emotion recognition in various fields including, but not limited to, human-computer interface \cite{hci}, animation \cite{aneja2016modeling}, medicine \cite{edwards2002emotion,chu2017facial} and security \cite{clavel2008fear,saste2017emotion}.

Emotion recognition can be performed using different features, such as face \cite{aneja2016modeling,mollahosseini2016going,liu2014facial}, speech \cite{han2014speech,clavel2008fear}, EEG \cite{petrantonakis2010emotion}, and even text \cite{wu2006emotion}. Among these features, facial expressions are one of the most popular, if not the most popular, due to a number of reasons; they are visible, they contain many useful features for emotion recognition, and it is easier to collect a large dataset of faces (than other means for human recognition) \cite{aneja2016modeling, lyons1998japanese, carrier2013fer}. 

\begin{figure}[t]
\begin{center}
    \begin{tabular}{cc}
    \includegraphics[width=0.68\linewidth]{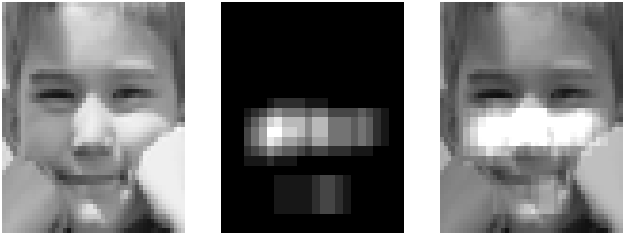} & \rotatebox{90}{\ \ \ \ Happiness} \\
    \includegraphics[width=0.68\linewidth]{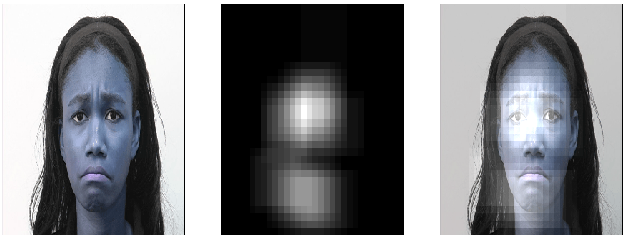} & \rotatebox{90}{\ \ \ \ \ Sadness} \\
    \includegraphics[width=0.68 \linewidth]{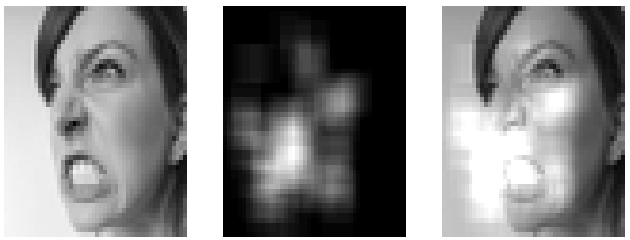} & \rotatebox{90}{\ \ \ \ \ Anger} \\
    \includegraphics[width=0.68 \linewidth]{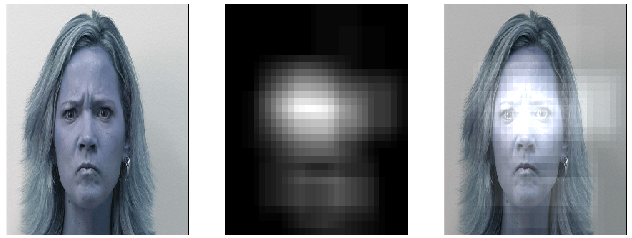} & \rotatebox{90}{\ \ \ \ \ Anger} \\
    \end{tabular}
\end{center}
   \caption{The detected salient regions for different facial expressions by our model. The images in the first and third rows are taken from FER dataset, and the images in the second and fourth rows belong to the extended Cohn-Kanade dataset. }
\label{fig:sample_res}
\vspace{-3mm}
\end{figure}
Recently, with the use of deep learning and especially convolutional neural networks (CNNs) \cite{lecun1989generalization}, many features can be extracted and learned for a decent facial expression recognition system \cite{khorrami2015deep,tzirakis2017end}. 
It is, however, noteworthy that in the case of facial expressions, much of the clues come from a few parts of the face, e.g. the mouth and eyes, whereas other parts, such as ears and hair, play little part in the output \cite{cohn1995computerized}. 
This means that ideally, the machine learning framework should  focus only on the important parts of the face,  and less sensitive to other face regions.

In this work we propose a deep learning based framework for facial expression recognition, which takes the above observation into account, and uses attention mechanism to focus on the salient part of the face. We show that by using attentional convolutional network, even a network with few layers (less than 10 layers) is able to achieve very high accuracy rate. 
More specifically, this paper presents the following contributions:
\begin{itemize}
\item We propose an approach based on an attentional convolutional network, which can  focus on feature-rich parts of the face, and yet, outperform remarkable recent works in accuracy.
\item In addition, we use the visualization technique proposed in \cite{fergus} to highlight the face image's most salient regions, i.e. the parts of the image which have the strongest impact on the classifier's outcome. Samples of salient regions for different emotions are shown in Figure \ref{fig:sample_res}.
\end{itemize}
In the following sections, we first provide an overview of related works in Section II.
The proposed framework and model architecture are explained in Section III. 
We will then provide the experimental results, overview of databases used in this work, and also model visualization in Section IV.
Finally we conclude the paper in Section V.

\section{Related Works}
\begin{figure}[h]
\begin{center}
   \includegraphics[page=2,width=\linewidth,trim={0cm 0.8cm 0cm 0cm}]{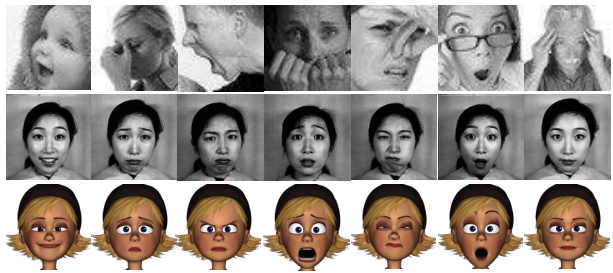}
\end{center}
   \caption{(Left to right) The six cardinal emotions (happiness, sadness, anger, fear, disgust, surprise) and neutral. The images in the first, second and the third rows belong to FER, JAFFE and FERG datasets respectively.}
\label{fig:FACSImage}
\end{figure}

In one of the most iconic works in emotion recognition by Paul Ekman \cite{ekman1971constants}, happiness, sadness, anger, surprise, fear and disgust were identified as the six principal emotions (besides neutral). 
Ekman later developed FACS \cite{FACS} using this concept, thus setting the standard for works on emotion recognition ever since. 
Neutral was also included later on, in most of human recognition datasets, resulting in seven basic emotions.
Image samples of these emotions from three datasets are displayed in Figure \ref{fig:FACSImage}.

Earlier works on emotion recognition, rely on the traditional two-step machine learning approach, where in the first step, some features are extracted from the images, and in the second step, a classifier (such as SVM, neural network, or random forest) are used to detect the emotions.
Some of the popular hand-crafted features used for facial expression recognition include the histogram of oriented gradients (HOG) \cite{mcconnell1986method,chen2014facial}, local binary patterns (LBP) \cite{shan2009facial}, Gabor wavelets \cite{bartlett2005recognizing} and Haar features \cite{whitehill2006haar}. 
A classifier would then assign the best emotion to the image. 
These approaches seemed to work fine on simpler datasets, but with the advent of more challenging datasets (which have more intra-class variation), they started to show their limitation. 
To get a better sense of some of the possible challenges with the images, we refer the readers to the images in the first row of Figure 2, where the image can have partial face, or the face can be occluded with hand or eye-glasses.

With the great success of deep learning, and more specifically convolutional neural networks for image classification and other vision problems \cite{alexnet,resnet,semseg,maskrcnn,iriscnn,adnet,aae,gan}, several groups developed deep learning-based models for facial expression recognition (FER).
To name some of the promising works, Khorrami in \cite{khorrami2015deep} showed that CNNs can achieve a high accuracy in emotion recognition and used a zero-bias CNN on the extended Cohn-Kanade dataset (CK+) and the Toronto Face Dataset (TFD)  to achieve state-of-the-art results.
Aneja et al \cite{aneja2016modeling} developed a model of facial expressions for stylized animated characters based on deep learning by training a network for modeling the expression of human faces, one for that of animated faces, and one to map human images into animated ones.
Mollahosseini \cite{mollahosseini2016going} proposed a neural network for FER using two convolution layers, one max pooling layer, and four ``inception'' layers, i.e. sub-networks.
Liu in \cite{liu2014facial} combines the feature extraction and classification in a single looped network, citing the two parts' need for feedback from each other. They used their Boosted Deep Belief Network (BDBN) on CK+ and JAFFE, achieving state-of-the-art accuracy.
Barsoum et al \cite{barsoum2016training} worked on using a deep CNN on noisy labels acquired via crowd-sourcing for ground truth images. They used 10 taggers to re-label each image in the dataset, and used various cost functions for their DCNN, achieving decent accuracy.
Han et al \cite{IBCNN} proposed an incremental boosting CNN (IB-CNN) in order to improve the recognition of spontaneous facial expressions by boosting the discriminative neurons, improving over the best methods of the time.
Meng in \cite{IACNN} proposed an identity-aware CNN (IA-CNN) which used identity- and expression-sensitive contrastive losses to reduce the variations in learning  identity- and expression-related information.


All of the above works achieve significant improvements over the traditional works on emotion recognition, but there seems to be missing a simple piece for attending to the important face regions for emotion detection. 
In this work, we try to address this problem, by proposing a framework based on attentional convolutional network, which is able to focus on salient face regions.


\begin{figure*}[t]
\begin{center}
   \includegraphics[page=1,width=\linewidth,trim={0cm 0.7cm 0cm 0cm}]{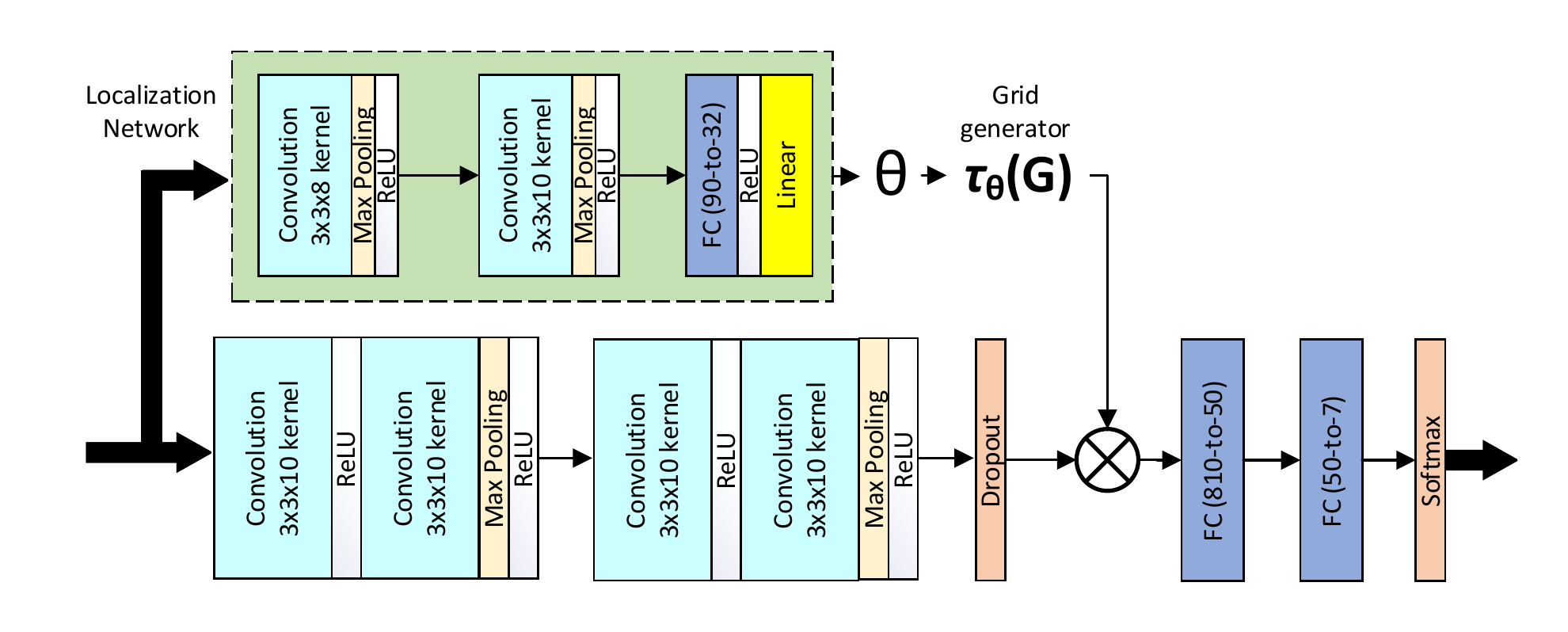}
\end{center}
   \caption{The proposed model architecture}
\label{fig:NetArch}
\end{figure*}

\section{The Proposed Framework}
We propose an end-to-end deep learning framework, based on attentional convolutional network, to classify the underlying emotion in the face images.
Often times, improving a deep neural network relies on adding more layers/neurons, facilitating gradient flow in the network (e.g. by adding adding skip layers), or better regularizations (e.g. spectral normalization), especially for classification problems with a large number of classes.
However, for facial expression recognition, due to the small number of classes, we show that using a convolutional network with less than 10 layers and attention (which is trained from scratch) is able to achieve promising results, beating state-of-the-art models in several databases.

Given a face image, it is clear that not all parts of the face are important in detecting a specific emotion, and in many cases, we only need to attend to the specific regions to get a sense of the underlying emotion.
Based on this observation, we add an attention mechanism, through spatial transformer network into our framework to focus on important face regions. 

Figure \ref{fig:NetArch} illustrates the proposed model architecture.
The feature extraction part consists of four convolutional layers, each two followed by max-pooling layer and rectified linear unit (ReLU) activation function. 
They are then followed by a dropout layer and two fully-connected layers.
The spatial transformer (the localization network) consists of two convolution layers (each followed by max-pooling and ReLU), and two fully-connected layers.
After regressing the transformation parameters, the input is transformed to the sampling grid $T(\theta)$ producing the warped data.
The spatial transformer module essentially tries to focus on the most relevant part of the image, by estimating a sample over the attended region. One can use different transformations to warp the input to the output, here we used an affine transformation which is commonly used for many applications.
For further details about the spatial transformer network, please refer to \cite{STN}

This model is then trained by optimizing a loss function using  stochastic gradient descent approach (more specifically Adam optimizer). The loss function in this work is simply the summation of two terms, the classification loss (cross-entropy), and the regularization term (which is $\ell_2$ norm of the weights in the last two fully-connected layers.

\begin{equation}
\mathcal{L}_{overall}= \mathcal{L}_{classifier}+ \lambda \|w_{(fc)}\|_2^2
\end{equation}

The regularization weight, $\lambda$, is tuned on the validation set. 
Adding both dropout and $\ell_2$ regularization enables us to train our models from scratch even on very small datasets, such as JAFFE and CK+. 
It is worth mentioning that we train a separate model for each one of the databases used in this work.
We also tried a network architecture with more than 50 layers, but the accuracy did not improve much. Therefore the simpler model shown here was used in the end.

\section{Experimental Results}
In this section we provide the detailed experimental analysis of our model on several facial expression recognition databases. We first provide a brief overview the databases used in this work, we then provide the performance of our models on four databases and compare the results with some of the promising recent works. 
We then provide the salient regions detected by our trained model using a visualization technique. 
\subsection{Databases}
In this work, we provide the experimental analysis of the proposed model on several popular facial expression recognition datasets, including FER2013 \cite{carrier2013fer}, the extended Cohn-Kanade \cite{ckplus}, Japanese Female Facial Expression (JAFFE) \cite{lyons1998japanese}, and Facial Expression Research Group Database (FERG) \cite{aneja2016modeling}.
Before diving into the results, we are going to give a brief overview of these databases.

\begin{figure}[h]
\begin{center}
   \includegraphics[width=1.02\linewidth]{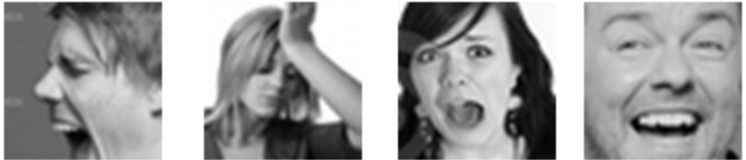}
\end{center}
   \caption{Four sample images from FER database}
\label{fig:FER}
\end{figure}

\textbf{FER2013}: The Facial Expression Recognition 2013 (FER2013) database was first introduced in the ICML 2013 Challenges in Representation Learning \cite{carrier2013fer}.
This dataset contains 35,887 images of 48x48 resolution, most of which are taken in wild settings. Originally the training set contained 28,709 images,and validation and test each include 3,589 images.
This database was created using the Google image search API and faces are automatically registered.
Faces are labeled as any of the six cardinal expressions as well as neutral.
Compared to the other datasets, FER has more variation in the images, including face occlusion (mostly with hand), partial faces, low-contrast images, and eyeglasses.
Four sample images from FER dataset are shown in Figure \ref{fig:FER}.

\begin{figure}[h]
\begin{center}
   \includegraphics[width=0.9\linewidth]{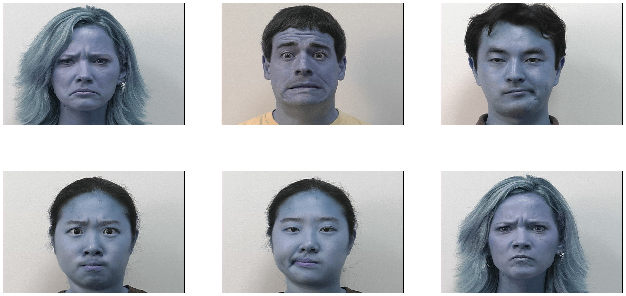}
\end{center}
   \caption{Six sample images from CK+ database}
\label{fig:CKdata}
\end{figure}

\textbf{CK+}:
The extended Cohn-Kanade (known as CK+) facial expression database \cite{ckplus} is a public dataset for action unit and emotion recognition.
It includes both posed and non-posed (spontaneous) expressions.
The CK+ comprises a total of 593 sequences across 123 subjects.
In most of previous works, the last frame of these sequences are taken and used for image based facial expression recognition.
Six sample images from this dataset are shown in Figure \ref{fig:CKdata}.

\begin{figure}[h]
\begin{center}
   \includegraphics[width=0.85\linewidth]{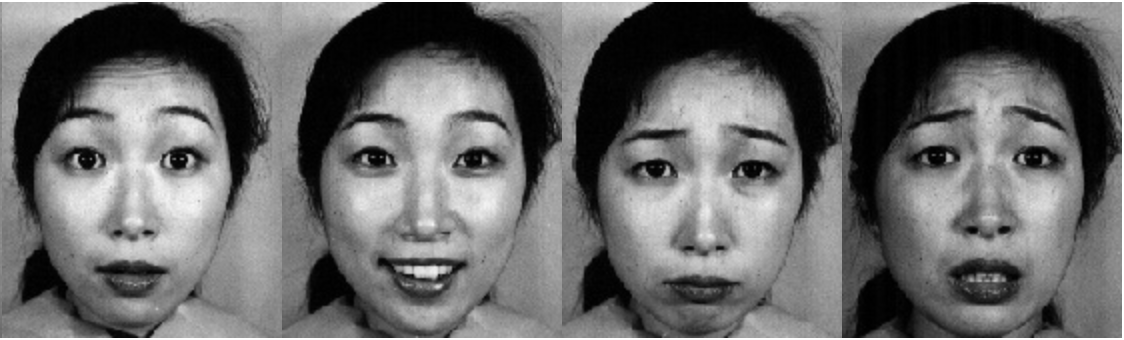}
\end{center}
   \caption{Four sample images from JAFFE database}
\label{fig:JAFFE}
\end{figure}

\textbf{JAFFE}: This dataset contains 213 images of the 7 facial expressions posed by 10 Japanese female models. 
Each image has been rated on 6 emotion adjectives by 60 Japanese subjects \cite{lyons1998japanese}. 
Four sample images from this dataset are shown in Figure \ref{fig:JAFFE}.

\begin{figure}[h]
\begin{center}
   \includegraphics[width=0.7\linewidth]{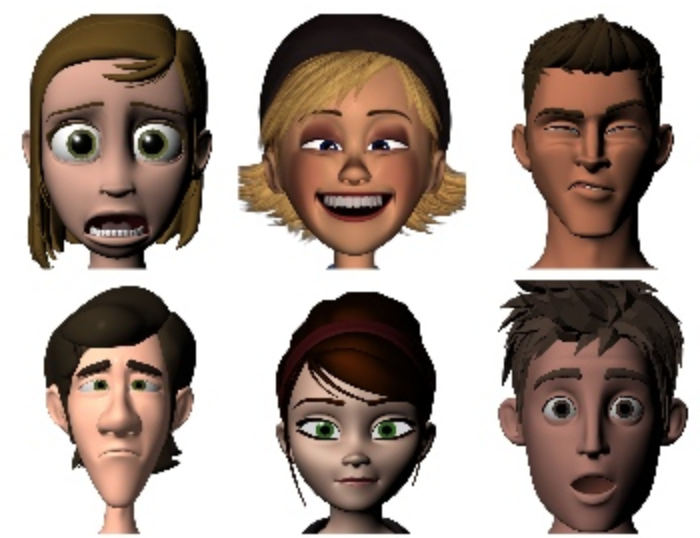}
\end{center}
   \caption{Six sample images from FERG database}
\label{fig:FERG}
\end{figure}

\textbf{FERG}:
FERG is a database of stylized characters with annotated facial expressions. The database contains 55,767 annotated face images of six stylized characters. The characters were modeled using MAYA. 
The images for each character are grouped into seven types of expressions \cite{aneja2016modeling}. Six sample images from this database are shown in Figure \ref{fig:FERG}.
We mainly wanted to try our algorithm on this database to see how it performs on cartoonish characters.

\subsection{Experimental Analysis and Comparison}
We will now present the performance of the proposed model on the above datasets. In each case, we train the model on a subset of that dataset and validate on validation set, and report the accuracy over the test set. 

Before getting into the details of the model's performance on different datasets, we briefly discuss our training procedure.
We trained one model per dataset in our experiments, but we tried to keep the architecture and hyper-parameters similar among these different models. 
Each model is trained for 500 epochs from scratch, on an AWS EC2 instance with a Nvidia Tesla K80 GPU.
We initialize the network weights with random Gaussian variables with zero mean and 0.05 standard deviation. 
For optimization, we used Adam optimizer with a learning rate of 0.005 with weight decay (Different optimizer were tried, including stochastic gradient descents, and Adam seemed to be performing slightly better).
It takes around 2-4 hours to train our models on FER and FERG datasets. For JAFFE and CK+, since there are much fewer images, it takes less than 10 minutes to train a model.
Data augmentation is used for the images in the training sets to train the model on a larger number of images, and make the trained model for invariant on small transformations.

\begin{figure}[h]
\begin{center}
   \includegraphics[width=0.9\linewidth]{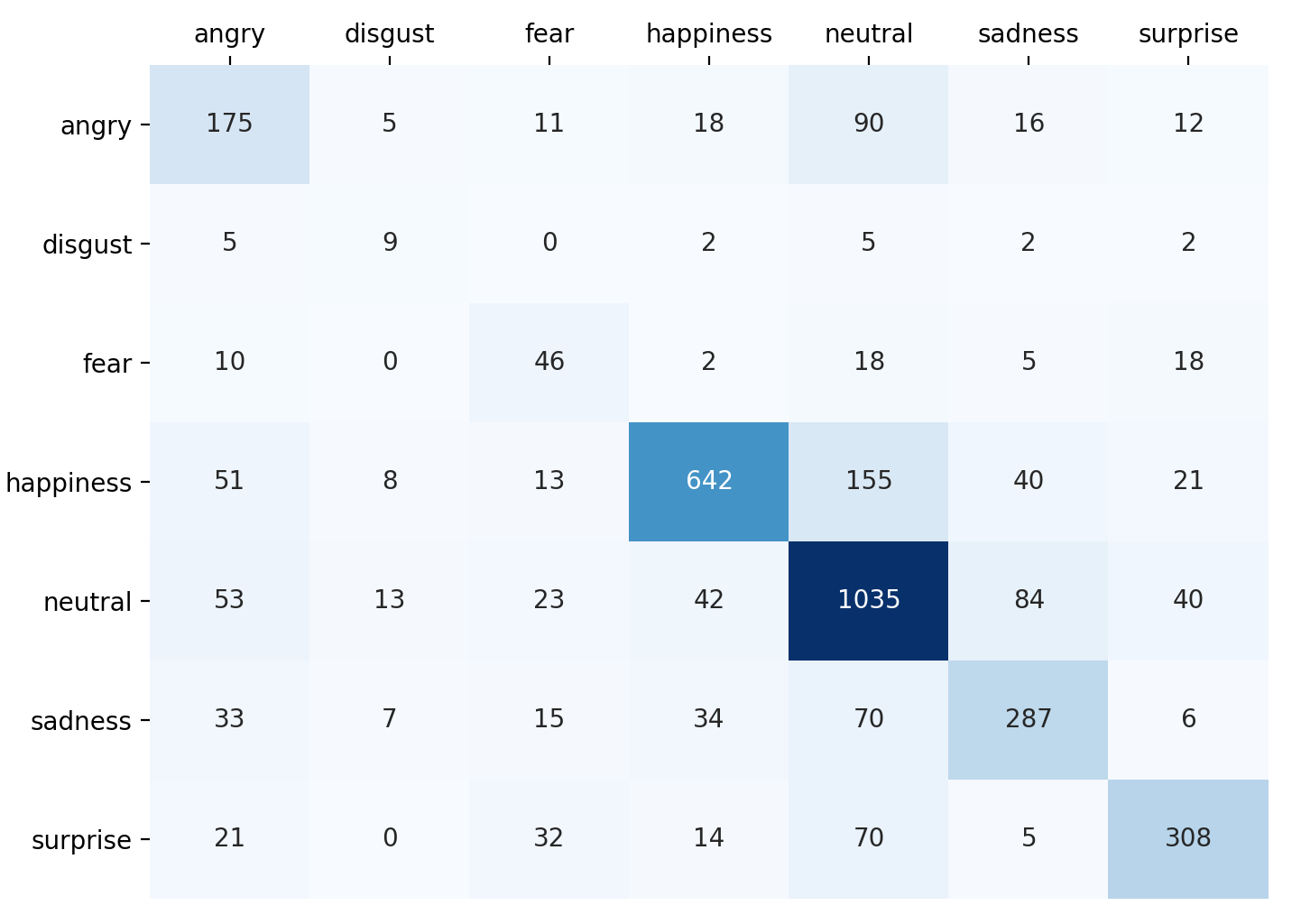}
\end{center}
   \caption{The confusion matrix of the proposed model on the test set of FER dataset}
\label{fig:ConfusionFER}
\end{figure}

As discussed before, FER-2013 dataset is more challenging than other facial expression recognition datasets we used. 
Besides the intra-class variation of FER, another main challenge in this dataset is the imbalance nature of different emotion classes. Some of the classes such as happiness and neutral have a lot more examples than others. 
We used the entire 28,709 images in the training set to train the model, validated on 3.5k validation images, and report the model accuracy on the 3,589 images in the test set. 
We were able to achieve an accuracy rate of around 70.02\% on the test set.
The confusion matrix on the test set of FER dataset is shown in Figure \ref{fig:ConfusionFER}. 
As we can see, the model is making more mistakes for classes with less samples such as disgust and fear.

The comparison of the result of our model with some of the previous works on FER 2013 are provided in Table \ref{table:FER}.
\begin{table}[h]
\centering
  \caption{Classification Accuracies on FER 2013 dataset}
\begin{tabular}{|m{4cm}|m{2cm}|}
\hline
Method  & Accuracy Rate\\
\hline
Bag of Words \cite{bow}  &  \ \ \ 67.4\%\\
\hline 
VGG+SVM \cite{vggsvm} &   \ \ \  66.31\% \\
\hline 
GoogleNet \cite{googlenet} &   \ \ \  65.2\% \\
\hline 
Mollahosseini et al \cite{mollahosseini2016going} &   \ \ \  66.4\% \\
\hline
 The proposed algorithm  &  \ \ \ 70.02\%\\
\hline
\end{tabular}
\label{table:FER}
\end{table}

\begin{figure}[t]
\begin{center}
   \includegraphics[width=0.99\linewidth]{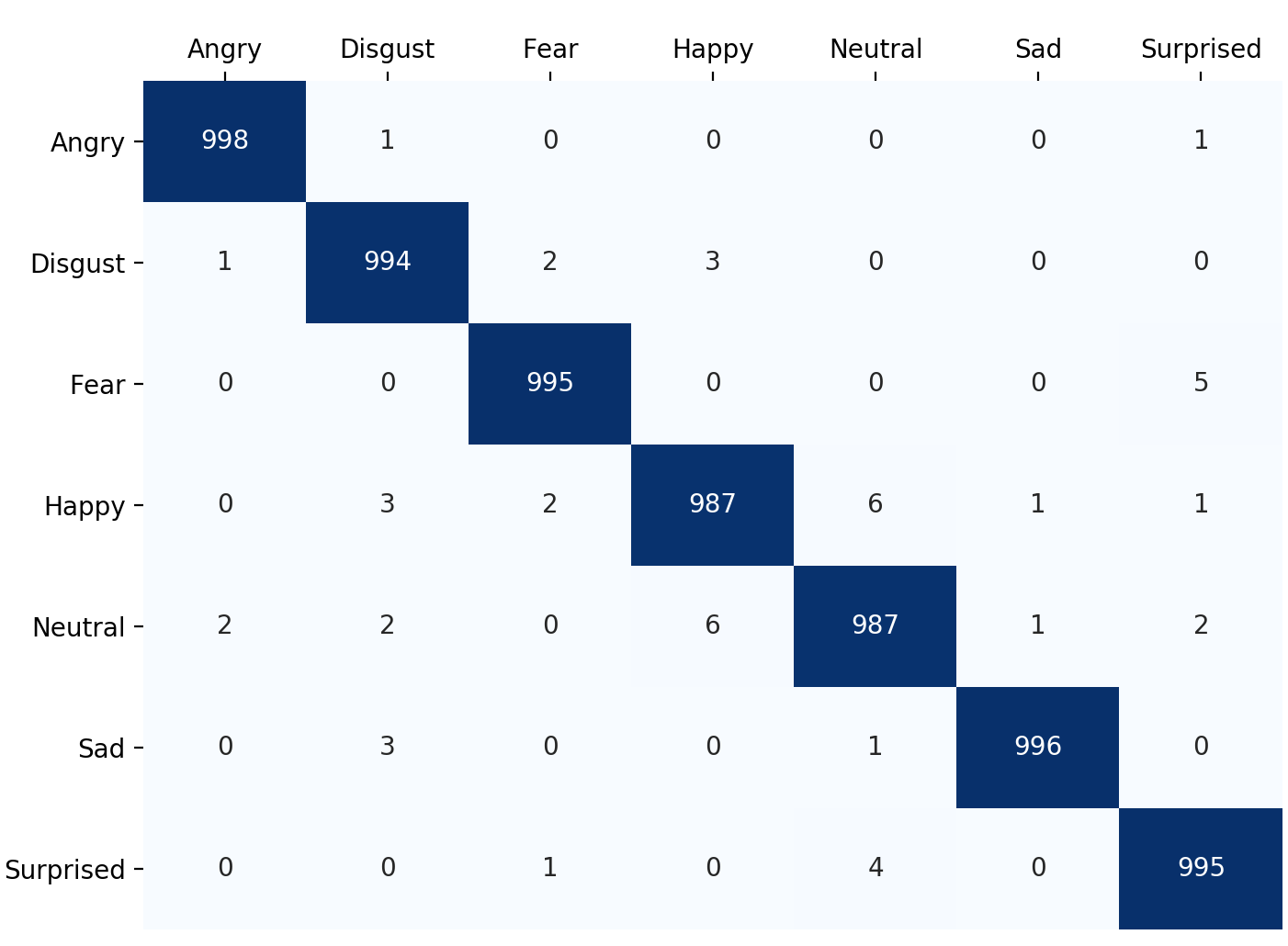}
\end{center}
   \caption{The confusion matrix on FERG dataset}
\label{fig:ConfusionFERG}
\end{figure}

For FERG dataset, we use around 34k images for training, 14k for validation, and 7k for testing. For each facial expression, we randomly select 1k images for testing. We were able to achieve an accuracy rate of around 99.3\%.
The confusion matrix on the test set of FERG dataset is shown in Figure \ref{fig:ConfusionFERG}.

The comparison between the proposed algorithm and some of the previous works on FERG dataset are provided in Table \ref{table:FERG}.
\begin{table}[h]
\centering
  \caption{Classification Accuracy on FERG dataset}
\begin{tabular}{|m{5cm}|c|}
\hline
Method  & Accuracy Rate\\
\hline
DeepExpr \cite{aneja2016modeling} & \ \ \ 89.02\% \\
\hline 
Ensemble Multi-feature \cite{ensemble} & \ \ \ 97\% \\
\hline 
Adversarial NN \cite{Ad-NN} &   \ \ \ \ 98.2\% \\
\hline
 The proposed algorithm  &  \ \ \ \ 99.3\%\\
\hline
\end{tabular}
\label{table:FERG}
\end{table}

\begin{figure}[t]
\begin{center}
   \includegraphics[width=0.85\linewidth]{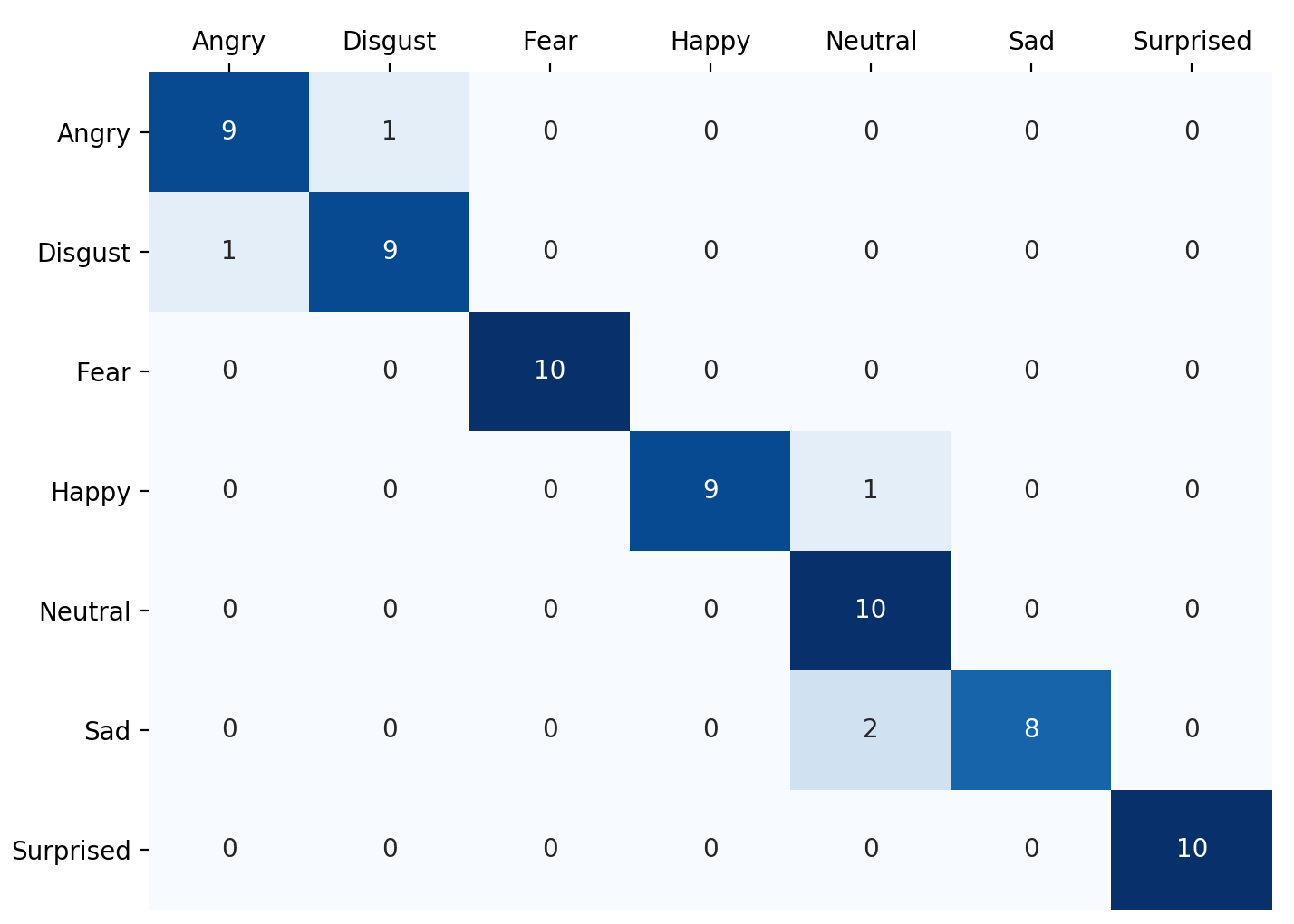}
\end{center}
   \caption{The confusion matrix on JAFFE dataset}
\label{fig:ConfusionJAFFE}
\end{figure}

For JAFFE dataset, we use 120 images for training, 23 images for validation, and 70 images for test (10 images per emotion in the test set). 
The confusion matrix of the predicted results on this dataset is shown in Figure \ref{fig:ConfusionJAFFE}. The overall accuracy on this dataset is around 92.8\%.

The comparison with previous works on JAFFE dataset are shown in Table \ref{table:JAFFE}.
\begin{table}[h]
\centering
  \caption{Classification Accuracy on JAFFE dataset}
\begin{tabular}{| m{5cm} | c |}
\hline
Method  & Accuracy Rate\\
\hline
Fisherface \cite{fisherface}  &  \ 89.2\%\\
\hline 
Salient Facial Patch \cite{SFP} &   \  91.8\% \\
\hline
CNN+SVM \cite{shima2018image} & 95.31\% \\
\hline
 The proposed algorithm  &  \ 92.8\%\\
\hline
\end{tabular}
\label{table:JAFFE}
\end{table}

For CK+, 70\% of the images are used as training, 10\% for validation, and 20\% for testing.
The comparison of our model with previous works onthe extended CK dataset are shown in Table \ref{table:CK}.
\begin{table}[h]
\centering
  \caption{Classification Accuracy on CK+}
\begin{tabular}{|m{4cm}|c|}
\hline
Method  & Accuracy Rate\\
\hline
MSR \cite{MSR}   &  \ 91.4\%\\
\hline 
3DCNN-DAP \cite{3DCNN} &   \  92.4\% \\
\hline 
Inception \cite{mollahosseini2016going} &   \  93.2\% \\
\hline 
IB-CNN \cite{IBCNN} &   \  95.1\% \\
\hline 
IACNN \cite{IACNN} &   \  \ 95.37\% \\
\hline
DTAGN \cite{DTAGN} &   \  97.2\% \\
\hline
ST-RNN \cite{STRNN} &   \  97.2\% \\
\hline
PPDN \cite{PPDN} & \ 97.3\% \\
\hline
 The proposed algorithm  &  \ 98.0\%\\
\hline
\end{tabular}
\label{table:CK}
\end{table}

\subsection{Model Visualization}
Here we provide a simple approach to visualize the important regions while classifying different facial expression, inspired by the work in \cite{fergus}. 
We start from the top-left corner of an image, and each time zero out a square region of size $N$x$N$ inside the image, and make a prediction using the trained model on the occluded image.
If occluding that region makes the model to make a wrong prediction on facial expression label, that region would be considered as a potential region of importance in classifying the specific expression. 
On the other hand, if removing that region would not impact the model's prediction, we infer that region is not very important in detecting the corresponding facial expression.
Now if we repeat this procedure for different sliding windows of $N$x$N$, each time shifting them with a stride of $s$, we can get a saliency map for the most important regions in detecting an emotion from different images.

We show nine example cluttered images for a happy and an angry image from JAFFE dataset, and how zeroing out different regions would impact the model prediction. 
As we can see, for the happy face zeroing out the areas around mouth would cause the model to make a wrong prediction, whereas for angry face, zeroing out the areas around eye and eyebrow makes the model to make a mistake.
\begin{figure}[h]
\begin{center}
   \includegraphics[width=0.98\linewidth]{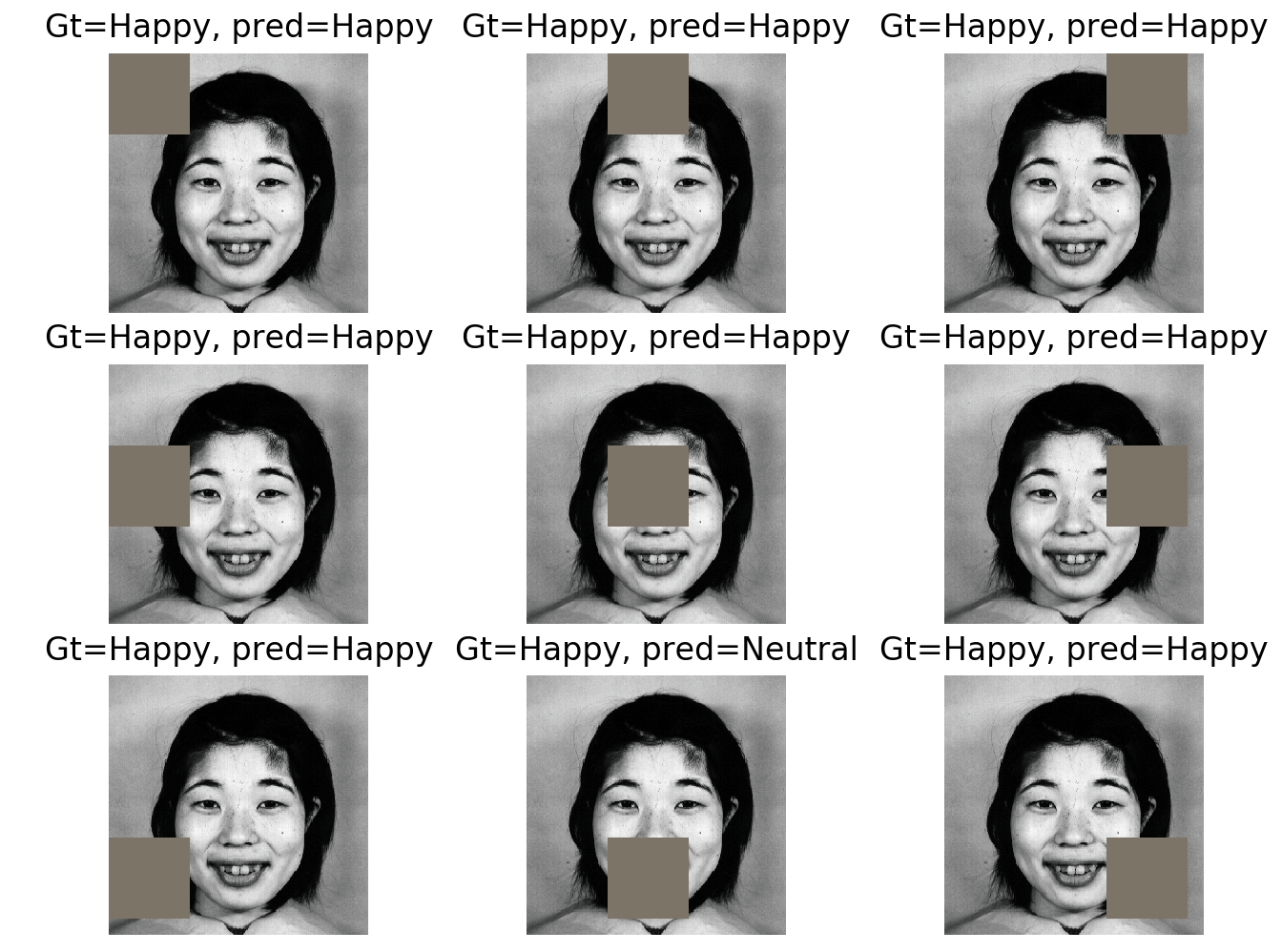}
   \includegraphics[width=0.95\linewidth]{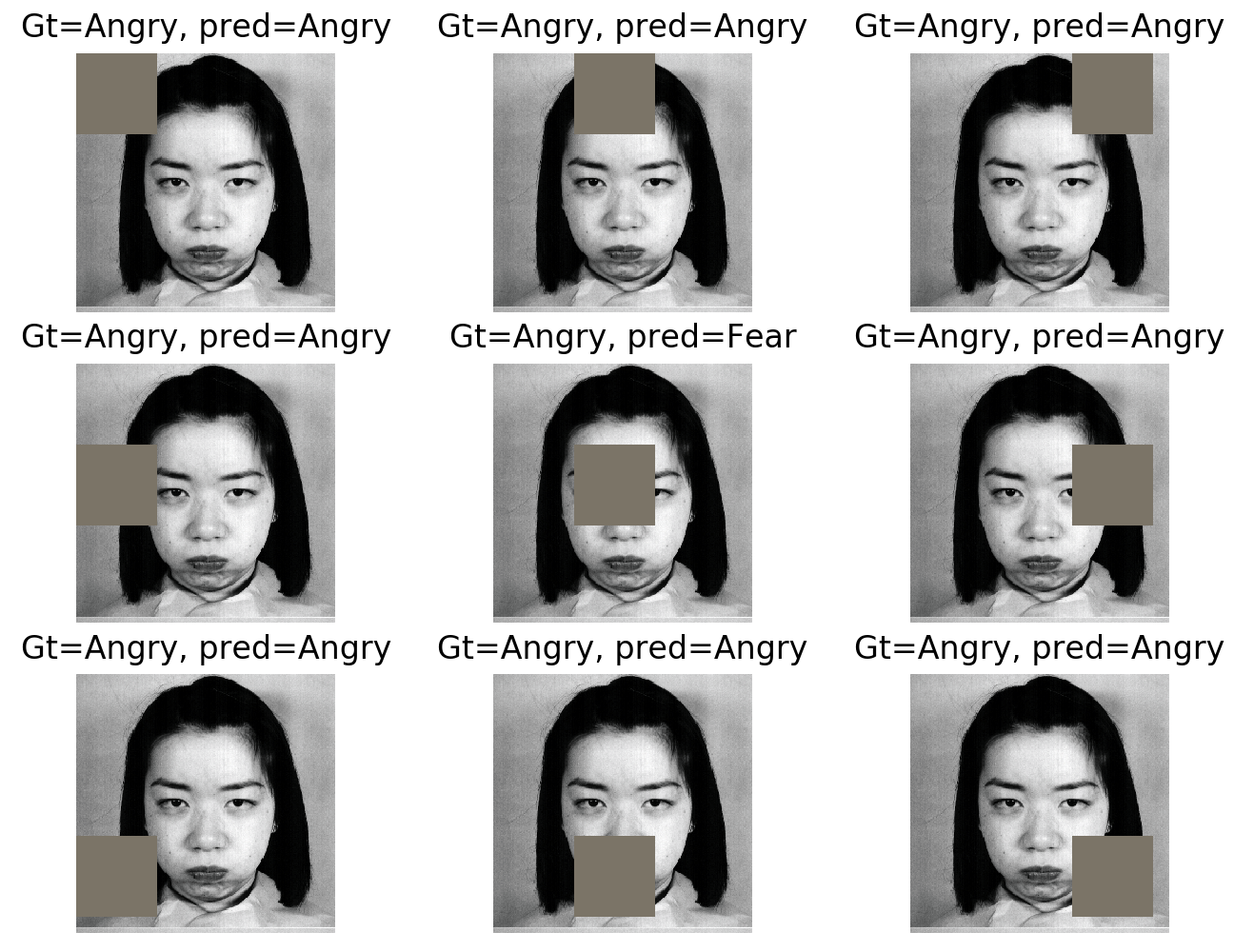}
\end{center}
   \caption{The impact of zeroing out different image parts, on the model prediction for a happy face (the top three rows) and an angry face (the bottom three rows).}
\label{fig:shifting}
\end{figure}

\begin{figure}[H]
\begin{center}
   \includegraphics[width=0.8\linewidth]{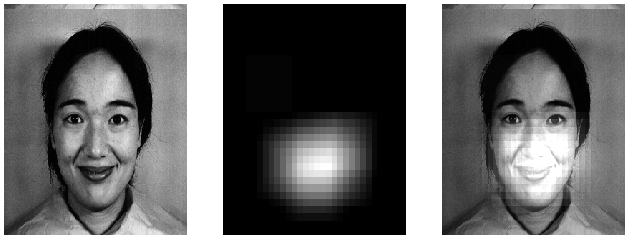} \rotatebox{90}{\ \ \ \ \ \ \  Happy} \\ \vspace{0.05cm}
   \includegraphics[width=0.8\linewidth]{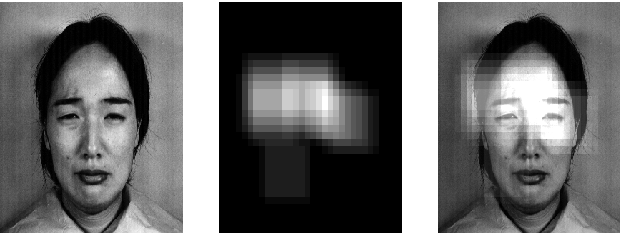} \rotatebox{90}{\ \ \ \ \ \ \   Sad} \\ \vspace{0.05cm}
   \includegraphics[width=0.8\linewidth]{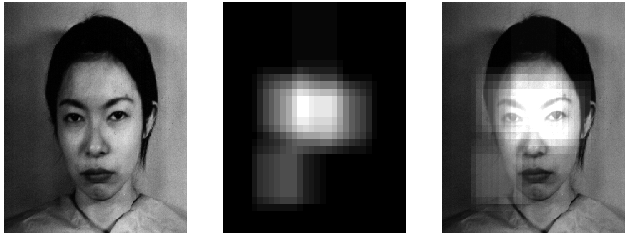} \rotatebox{90}{\ \ \ \ \ \ \   Angry} \\ \vspace{0.06cm}
   \includegraphics[width=0.8\linewidth]{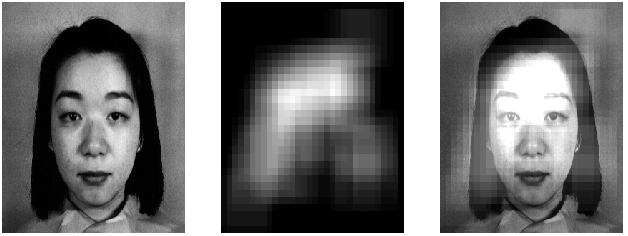} \rotatebox{90}{\ \ \ \ \ \ \   Neutral} \\ \vspace{0.09cm}
    \includegraphics[width=0.8\linewidth]{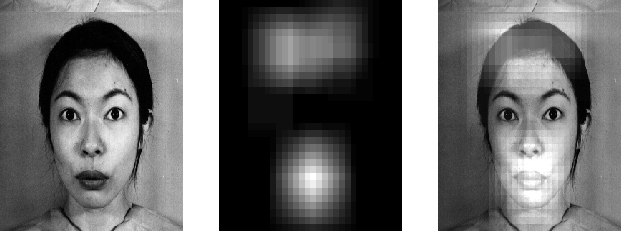} \rotatebox{90}{\ \ \ \ \  Surprised} \\ \vspace{0.1cm}
    \includegraphics[width=0.8\linewidth]{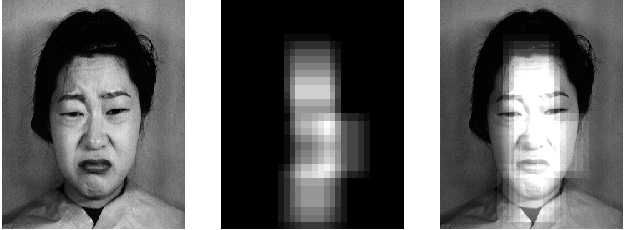} \rotatebox{90}{\ \ \ \ \ \   Disgust} \\ \vspace{0.12cm}
    \includegraphics[width=0.8\linewidth]{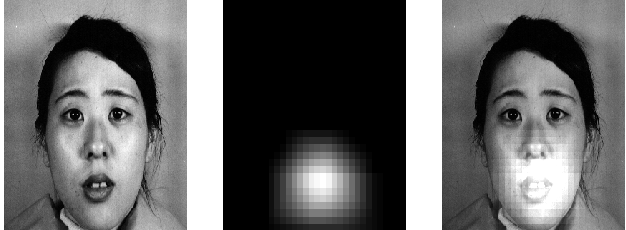} \rotatebox{90}{\ \ \ \ \ \ \  Fear} \\
\end{center}
   \caption{The important regions for detecting different facial expressions. As it can be seen, the saliency maps for happiness, fear and surprised are sparser than other emotions.}
\label{fig:SaliencyJAFFE}
\end{figure}

Figure \ref{fig:SaliencyJAFFE}, shows the important regions of 7 sample images from JAFFE dataset, each corresponding to a different emotion.
There are some interesting observations from these results. 
For example, for the sample image with neutral emotion in the fourth row, the saliency region essentially covers the entire face, which means that all these regions are important to infer that a given image has neutral facial expression. 
This makes sense, since changes in any parts of the face (such as eyes, lips, eyebrows and forehead) could lead to a different facial expression, and the algorithm needs to analyze all those parts in order to correctly classify a neutral image.
This is however not the case for most of the other emotions, such as happiness, and fear, where the areas around the mouth turns out to be more important than other regions.

\begin{figure}[h]
\begin{center}
   \includegraphics[width=0.6\linewidth]{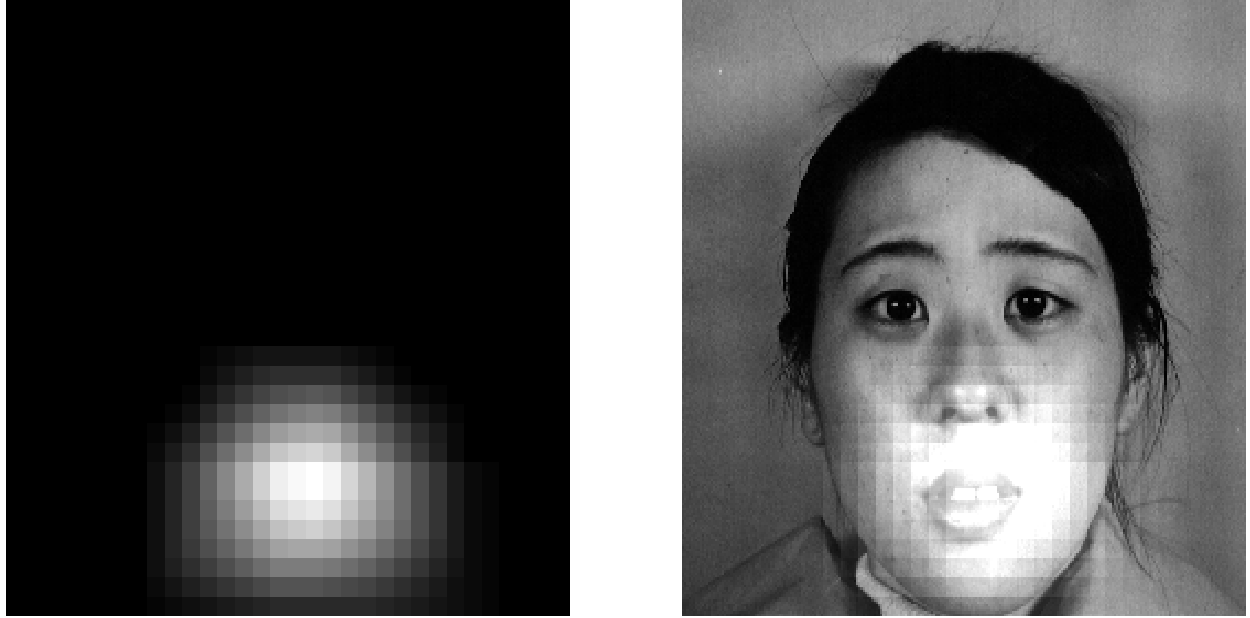}
   \includegraphics[width=0.6\linewidth]{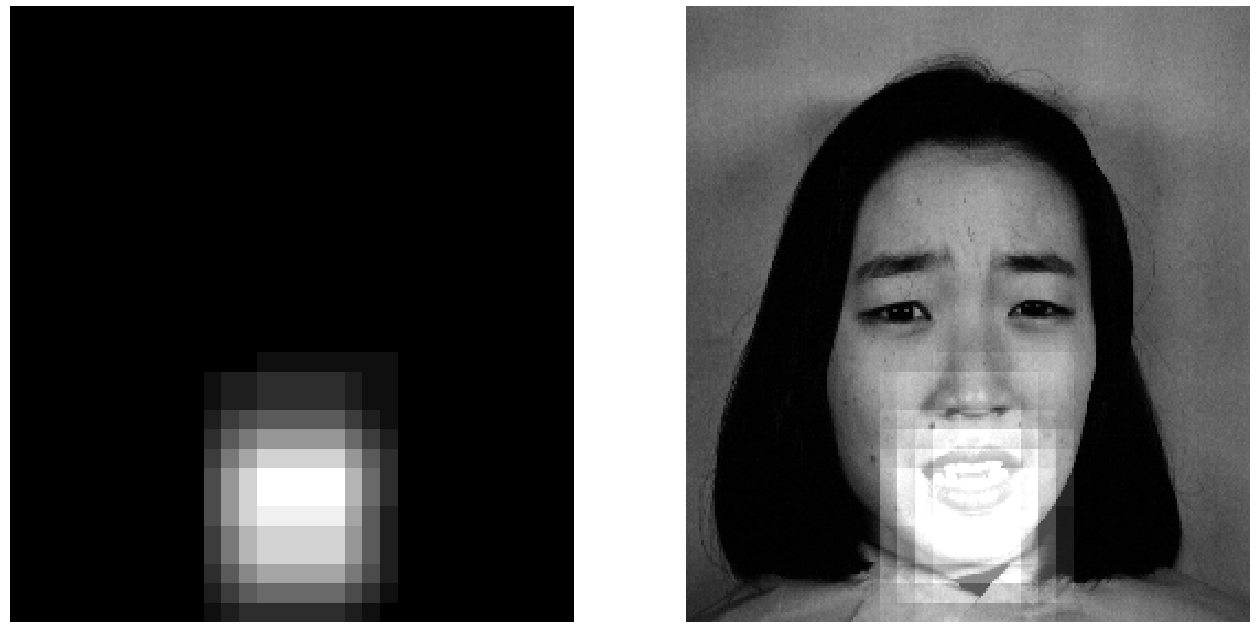}
   \includegraphics[width=0.6\linewidth]{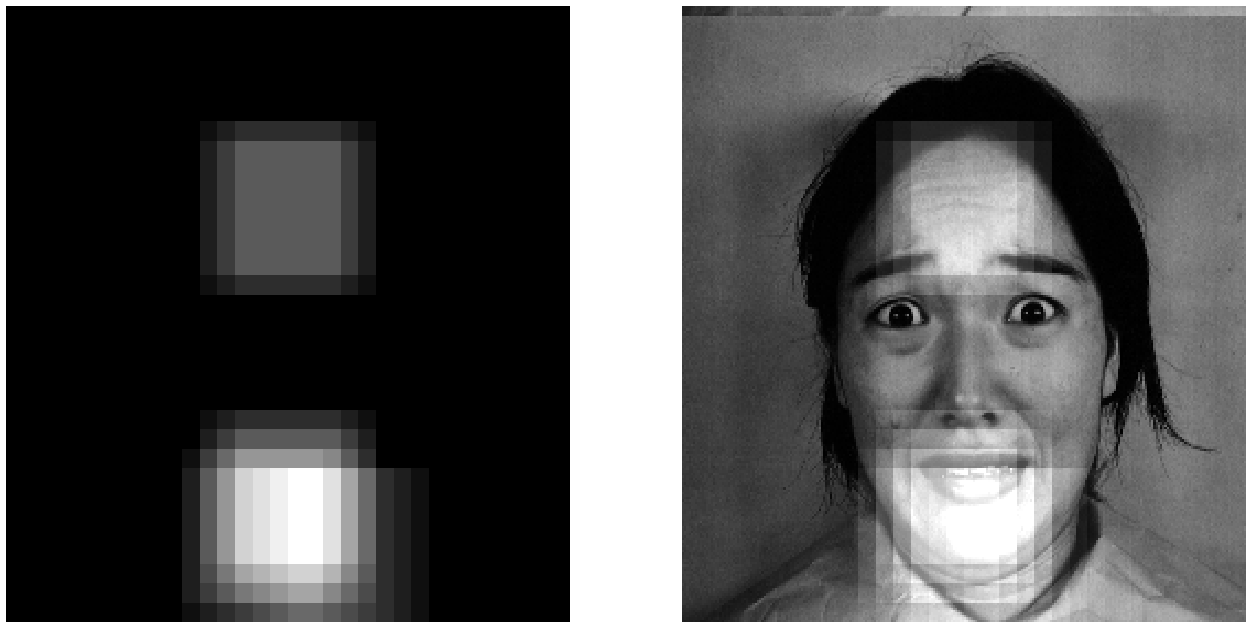}
\end{center}
   \caption{The important regions for three images with fear}
\label{fig:SaliencyFear}
\end{figure}

It is worth mentioning that different images with the same facial expression could have different saliency maps due to the different gestures and variations in the image. 
In Figure \ref{fig:SaliencyFear}, we show the important regions for three images with facial expression of ``fear''. 
As it can be seen from this figure, the important regions for these images are very similar in detecting the mouth, but the last one also considers some part of forehead in the important region. 
This could be because of the strong presence of forehead lines, which is not visible in the two other images.

\section{Conclusion}
This paper proposes a new framework for facial expression recognition using an attentional convolutional network.
We believe attention is an important piece for detecting facial expressions, which can enable neural networks with less than 10 layers to compete with (and even outperform) much deeper networks for emotion recognition.
We also provided an extensive experimental analysis of our work on four popular facial expression recognition databases, and showed promising results.
Also, we have deployed a visualization method to highlight the salient regions of face images which are the most crucial parts thereof in detecting different facial expressions.

\section*{Acknowledgment}
We would like to express our gratitude to the people at University of Washington Graphics and Imaging Lab (GRAIL) for providing us with access to the FERG database.
We would also like to thank our colleagues and partners for reviewing our work, and providing very useful comments and suggestions.


\end{document}